\documentclass[lettersize,journal]{IEEEtran}

\usepackage{amsmath,amsfonts}
\usepackage{amssymb}
\usepackage{mathtools}
\usepackage{amsthm}
\usepackage{microtype}
\usepackage{graphicx}
\usepackage{booktabs}
\usepackage{algorithm}
\usepackage{algorithmic}
\usepackage{array}
\usepackage[caption=false,font=normalsize,labelfont=sf,textfont=sf]{subfig}
\usepackage{textcomp}
\usepackage{stfloats}
\usepackage{url}
\usepackage{verbatim}
\usepackage{cite}
\usepackage{placeins}    
\usepackage{enumitem}
\usepackage[table]{xcolor}
\usepackage{multirow}
\usepackage[colorlinks=true,linkcolor=black,citecolor=black,urlcolor=black]{hyperref}
\usepackage[capitalize,noabbrev]{cleveref}

\theoremstyle{plain}

\theoremstyle{definition}

\theoremstyle{remark}

\usepackage[textsize=tiny]{todonotes}

\definecolor{HeaderGray}{RGB}{245,245,245}
\definecolor{BestBlue}{RGB}{0,114,178}
\definecolor{OursBlue}{RGB}{227,242,253}

\begin{document}

\title{TRIO: Token Reduction via Inference-Objective Guidance\\
for Efficient Vision-Language Models}

\author{Haokui Zhang, Congyang Ou, Dawei Yan, Peng Wang, Qingsen Yan,Yu Zhang, Ying Li,\textsuperscript{\ensuremath{\dagger}} and Rong Xiao%
\thanks{Haokui Zhang, Congyang Ou, and Dawei Yan are with the School of Cyberspace Security, Northwestern Polytechnical University, Xi'an, Shaanxi, China.}
\thanks{Peng Wang, Qingsen Yan, Yu Zhang, and Ying Li are with the School of Computer Science, Northwestern Polytechnical University, Xi'an, Shaanxi, China. (e-mail: \textless{}lybyp@nwpu.edu.cn\textgreater{}).}%
\thanks{Rong Xiao is with Intellifusion, China.}%
\thanks{*Corresponding author: Ying Li.}%
}

\markboth{IEEE Transactions on Multimedia}%
{}

\maketitle

\begin{abstract}

Recently, reducing redundant visual tokens in vision-language models (VLMs) to accelerate VLM inference has emerged as a hot topic. However, most existing methods rely on heuristics constructed based on inter-visual-token similarity or cross-modal visual-text similarity, which gives rise to certain limitations in compression performance and practical deployment. In contrast, we propose TRIO from the perspective of inference objectives, which transforms visual token compression into preserving output result invariance and selects tokens primarily by their importance to this goal. 
Specifically, vision tokens are reordered with  the guidance of token-level gradient saliency generated by our designed layer-local proxy loss, a coarse constraint from the current layer to the final result. 
Then the most valuable vision tokens are selected following the non-maximum suppression (NMS) principle.
The proposed TRIO is training-free and compatible with FlashAttention, friendly to practical application and deployment. It can be deployed independently as an encoder-free method, or combined with encoder compression approaches like VisionZip for use as an encoder-involved method. On LLaVA-Next-7B, TRIO retains just 11.1\% of visual tokens but maintains 97.2\% of the original performance, with a 2.75$\times$ prefill speedup, 2.14$\times$ inference speedup, 6.22$\times$ lower FLOPs, and 6.05$\times$ reduced KV Cache overhead.Our code is available at\url{https://github.com/ocy1/TRIO}. 
\end{abstract}


\section{Introduction}\label{Introduction}
\IEEEPARstart{B}{y} incorporating visual information, vision-language models (VLMs) extend the strong language understanding and reasoning abilities of large language models (LLMs) to multimodal scenarios, such as image understanding, visual question answering, and multimodal reasoning. Representative VLMs, including Flamingo~\cite{alayrac2022flamingo}, BLIP-2~\cite{li2023blip}, LLaVA~\cite{liu2023llava}, and Qwen-VL~\cite{bai2023qwen}, have achieved promising performance across diverse tasks. However, the introduced visual tokens substantially increase the input sequence length, leading to considerable self-attention computation and KV-cache memory overhead. Therefore, efficient VLM acceleration has become a key problem for practical deployment.

Recently, a growing body of work has focused on compressing redundant visual tokens to reduce the computational cost and memory overhead of VLMs. Based on whether the vision encoder is involved in the compression process, existing methods can be roughly divided into two categories: \emph{Vision Encoder-Involved (w/ VE)} and \emph{Vision Encoder-Free (w/o VE)}. The former performs token merging, selection, or reorganization on the vision encoder side, thereby shortening the visual sequence before it is fed into the LLM and directly reducing the subsequent attention computation. Representative methods include ToMe~\cite{bolya2023tome}, VisionZip~\cite{yang2025visionzip}, HoloV~\cite{holov}, and SCOPE~\cite{deng2025scope}. In contrast, the latter keeps the outputs of the vision encoder unchanged and selects or prunes visual tokens within the LLM according to their importance, which often makes such methods more plug-and-play. Representative works include FastV~\cite{chen2024FastV}, SparseVLM~\cite{zhang2025sparsevlm}, and PyramidDrop~\cite{xing2024pyramiddrop}. These two lines of research reduce visual-token redundancy at different stages and provide effective routes toward efficient VLM inference.

\begin{figure*}[t]
    \centering
    \includegraphics[width=\linewidth]{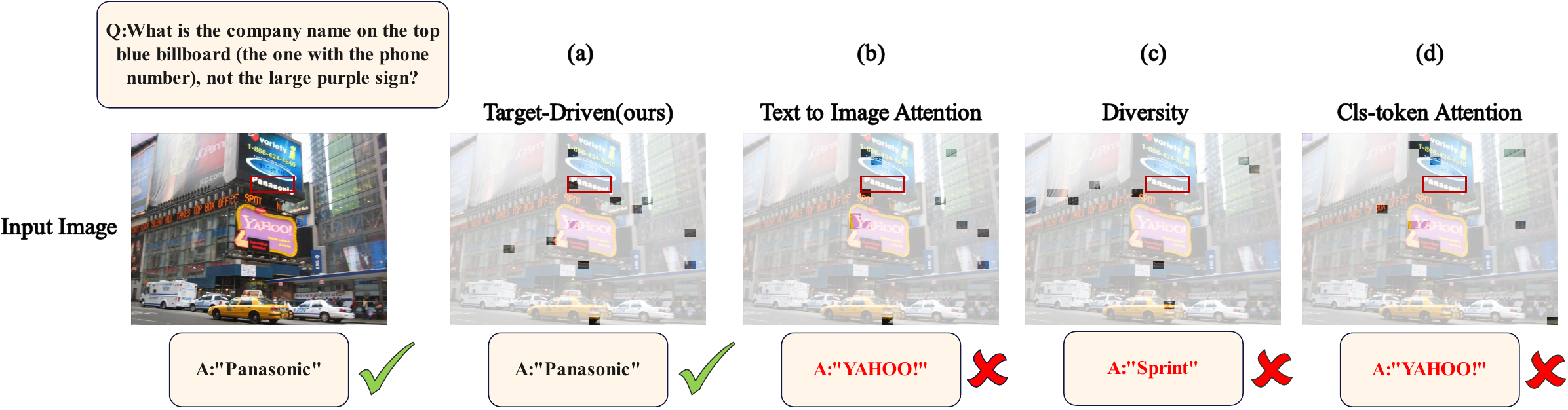}
    \caption{Comparison of different token selection strategies (selecting only 12 visual tokens as an example). (a) Ours; (b) Similirty with cls token and text token based; (c) cls token similarity based and enhance diversity; (d) Cls token similarity based.More examples are given in Appendix \ref{sec:vis_case_study_en}.}
    \label{fig:four-methods}
\end{figure*}

Despite the progress made by the above methods, they still suffer from several limitations. 
Current methods have two major core ideas: 1) Attention map-based compression. Redundant tokens are filtered out by the importance of each token relative to others after the attention map is calculated in the self-attention module. Yet attention map computation accounts for the main computational cost of LLMs, so existing models usually adopt the FlashAttention mechanism \cite{dao2022flashattention} for iterative replacement to address this issue. However, full attention map-based methods suffer from poor compatibility with FlashAttention, which impairs the actual acceleration effect. 2) Similarity-based compression. Specifically, filtering is conducted by measuring the similarity of visual tokens to the cls token (global token) (Fig.\ref{fig:four-methods}(d)); the similarity among the selected tokens is also taken into account to ensure their diversity (Fig.\ref{fig:four-methods}(c)); and a comprehensive metric is adopted by jointly considering the similarities of visual tokens to both the cls token and text tokens (Fig.\ref{fig:four-methods}(b)). As shown in Fig.\ref{fig:four-methods}, Mechanistically, all three of these methods carry the potential for misjudgment. Fig.\ref{fig:four-methods} (c)(d), The information encoded in the cls token is overly coarse and irrelevant to the input query, resulting in an incorrect final output. Fig.\ref{fig:four-methods}(b), when text tokens are taken into account, the phrase the large purple sign in the query interferes with the judgment, leading to a bias toward "YAHOO!" on the purple sign.

We conjecture that both attention map-based and similarity-based methods select and filter visual tokens according to pre-defined rules, which may fail to cover all cases. Here, we rethink the VLM acceleration problem from an inferential perspective. From the perspective of preserving reasoning result consistency, we propose a training-free, FlashAttention-compatible visual feature compression method that enables plug-and-play integration into practical deployment for inference acceleration. Specifically, we design a highly simple layer-local proxy loss to roughly estimate the impact of vision tokens at any layer on preserving the current output. Based on this loss function, we derive the saliency of each vision token and then reorder vision tokens according to saliency. Furthermore, inspired by the NMS algorithm, we design an NMS-based selection algorithm to perform final selection on the reordered tokens. This operation is implemented in the LLM component, executed across multiple layers in a shallow-to-deep fashion with an increasingly intensive screening strength.

In summary, our contributions are three-fold:

\begin{itemize}
    \item We present TRIO, a training-free visual token reduction method. To the best of our knowledge, this is the first goal-driven VLM acceleration method that does not rely solely on similarity or attention maps.
    \item Particularly, we design a layer-local proxy loss and an NMS selection method, both of which feature high computational efficiency and FlashAttention compatibility, enabling practical inference acceleration independent of code or hardware optimizations, as verified in Table \ref{tab:kpos_ablation_avg}.
    \item Experimental results on three popular VLMs and eight benchmarks verify the efficiency of TRIO.  
   
\end{itemize}

\section{Related Work}\label{Related Work}

\subsection{Multimodal Large Language Models.}

Most multimodal large language models (VLMs) follow a strong-LLM-centered paradigm, where a vision encoder extracts image or video features and a projection layer, resampler, or cross-attention module aligns visual representations with the language space. Representative early models include Flamingo~\cite{alayrac2022flamingo}, which connects frozen vision and language backbones via gated cross-attention for few-shot multimodal inference, and BLIP-2~\cite{li2023blip}, which introduces a Q-Former to efficiently bridge a frozen vision encoder and a frozen LLM. Later, MiniGPT-4~\cite{zhu2023minigpt}, InstructBLIP~\cite{dai2023instructblip}, and LLaVA~\cite{liu2023llava} demonstrate the effectiveness of visual instruction tuning for building general-purpose multimodal assistants, while Qwen-VL~\cite{bai2023qwen} further systematizes the architecture, training data, and task coverage of vision-language modeling. Meanwhile, KOSMOS-2~\cite{peng2023kosmos}, PaLM-E~\cite{driess2023palm}, InternVL~\cite{chen2024internvl}, and CogVLM~\cite{wang2024cogvlm} further advance general multimodal understanding, embodied reasoning, and open-source VLM capabilities. This paradigm has also been extended to video understanding, as represented by Video-ChatGPT~\cite{maaz2024video} and Video-LLaVA~\cite{lin2024video}. Gemini~\cite{team2023gemini} represents natively multimodal foundation models and shows strong capabilities across diverse multimodal tasks. However, as high-resolution images, multi-image inputs, and video understanding become increasingly common, the number of visual tokens grows rapidly, leading to substantial self-attention computation and KV-cache memory overhead. Although efficient attention implementations such as FlashAttention~\cite{dao2022flashattention} can alleviate the memory-access bottleneck in attention computation, they do not fundamentally remove the cost caused by the increasing visual sequence length. Therefore, visual token compression remains an important problem for efficient VLM inference.

\subsection{Visual Token Compression for VLMs.}
Existing visual token compression methods can be grouped by where they operate: \emph{Vision Encoder-Involved (w/ VE)} and \emph{Vision Encoder-Free (w/o VE)}. w/ VE methods compress visual tokens inside the vision encoder or before visual features are fed into the LLM. For example, ToMe~\cite{bolya2023tome} merges tokens by feature similarity, VisionZip~\cite{yang2025visionzip} selects representative tokens based on CLS similarity, and LLaVA-PruMerge~\cite{shang2025llava}, FasterVLM~\cite{zhang2024cls}, TokenPacker~\cite{li2025tokenpacker},and FastVLM~\cite{vasu2025fastvlm} reduce the number of visual tokens from the perspectives of visual redundancy, projector design, or vision encoder design. HoloV~\cite{holov}, SCOPE~\cite{deng2025scope}, DivPrune~\cite{alvar2025divprune}, and CDPruner~\cite{zhang2025cdpruner} further combine attention cues, spatial coverage, or content diversity for early compression. In contrast, w/o VE methods keep the outputs of the vision encoder unchanged and prune visual tokens inside the LLM. FastV~\cite{chen2024FastV}, SparseVLM~\cite{zhang2025sparsevlm}, and PyramidDrop~\cite{xing2024pyramiddrop} remove redundant visual tokens using text-to-image attention, key-text attention, or progressive pruning strategies, while VTW~\cite{lin2025boosting}, TopV~\cite{yang2025topv}, DART~\cite{wen2025dart}, PACT~\cite{dhouib2025pact}, FitPrune~\cite{ye2025fit} further reduce LLM-side overhead through deep-layer withdrawal, redundancy modeling, clustering-based merging, or graph-based token selection. For video scenarios, PruneVid~\cite{huang2025prunevid} explores visual token pruning under spatio-temporal redundancy. TRIO belongs to the w/o VE category, while it can also be combined with w/ VE methods such as VisionZip to form a joint compression scheme. Both modes are experimentally validated in this paper.


\begin{figure*}
  \centering
  \includegraphics[width=0.9\linewidth]{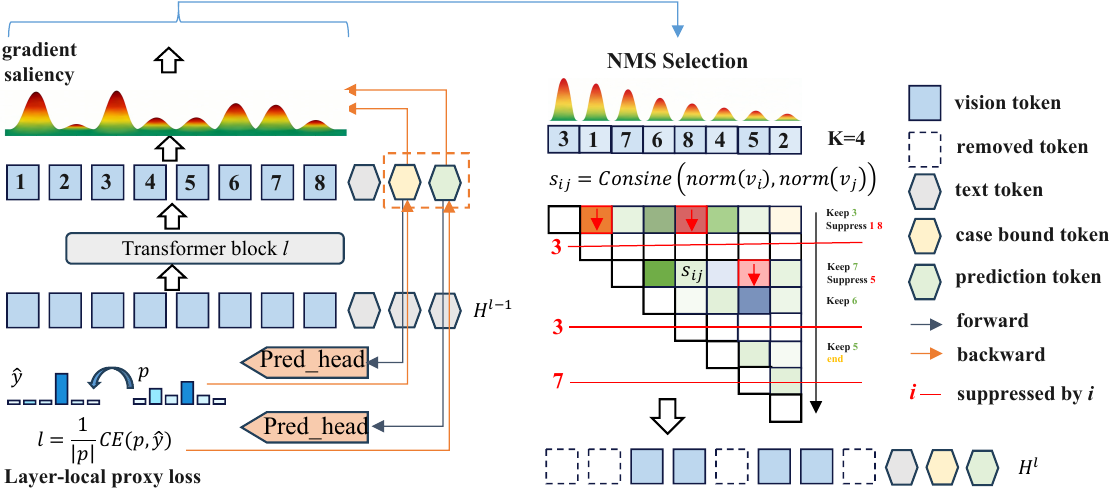}
  \caption{The architecture of TRIO. The overall framework consists of two stages. As shown on the left, the first stage is responsible for reordering vision tokens according to gradient saliency. The right side illustrates the second stage, which selects tokens based on the proposed NMS strategy.The $pred\_head$ is a pre-trained component native to the model, a prediction head originally designed to operate on the final-layer features.}
  \label{fig:overall}
\end{figure*}
\section{Method}\label{Method}

In this section, we explain the principles and detailed design of TRIO in detail. First, we introduce its underlying logic, and then we describe the specific architecture of the method.

We know that the gradient magnitude of each token reflects the importance of the token in driving the model toward the ground-truth output \cite{selvaraju2017grad}. As a straightforward extension, if we treat gradients as saliency to guide token selection and retain tokens with high saliency, we can basically ensure that the model output still converges to the ground truth. In other words, we derive the token selection strategy in a reverse manner from the perspective of guaranteeing output performance.

There are two key challenges in generating such a gradient signal: 1) how to construct a computationally efficient loss function; 2) what constitutes the ground truth for the loss function. It is infeasible to construct a precise loss function that meets the required criteria, as this would require inference to the final layer, thus negating the purpose of acceleration. As shown in the left side of Fig.\ref{fig:overall}, we propose the layer-local-proxy loss as a coarse approximation. Gradient-guided selection alone results in a lack of diversity in the selected tokens. Inspired by the NMS algorithm, we design an NMS selection algorithm to perform token selection based on gradient signal. We next elaborate on the design in the form of mathematical expressions.


\subsection{ Problem Formulation}

We consider a multimodal model with a vision encoder and an LLM decoder. Given an image $I$ and instruction $X$, the encoder outputs visual tokens $V=Enc(I)$, which are projected as $\tilde{V}=Proj(V)$ and concatenated with text embeddings $T$ to form the decoder input $H^{(0)}=[\tilde{V};T]$. The $L$-layer decoder updates.
\begin{equation}
H^{l} = f^{l}\!\left(H^{l-1}\right), \quad l=1,\ldots,L,
\end{equation}
where $l$ is the layer number, $f^{l}$ means the transformer block of the $l$-th layer. 
During the prefill stage, we select several pruning layers for compression from shallow to deep, with the number of retained tokens decreasing progressively. 


\subsection{Layer-local Proxy Loss and Gradient Saliency}\label{sec3.2}



At each pruning layer $l_{prune}$, we first compute the layer-local proxy loss using the following formula:

\begin{equation}
	L^{l} = \frac{1}{\lvert \mathcal{P}^{l} \rvert}
\sum_{t \in \mathcal{P}^{l}} \mathrm{CE}\!\left(p_t^{l}, \hat{y}_t^{l}\right),
\end{equation}
where $CE(\cdot, \cdot)$ denotes the standard cross-entropy loss. $\mathcal{P}^{l}$ is the set of indices for tokens used in loss calculation. $p_{t}^{l}$ and $\hat{y}_t^{l}$ are soft logit and hard logit, which are calculated with formulas:
\begin{equation}
\begin{aligned}
&Z^{l}=Head(Norm(H^{l} )), \\
&p_{t}^{l} =softmax(z_{t}), \\
&\hat{y}_t^{l} = \arg\max_{i} \, p_t^{l}(i),
\end{aligned}	
\end{equation}
where $Head$ denotes the original prediction head, used in the final layer of the original LLM. Here, we leverage it to generate soft logits. $z_{t}$ is the $t$-th token in $Z^{l}$. $Norm$ is the original layer normalization function of layer $l$. Here, we present two rough estimates as follows:
\begin{itemize}
    \item Using shallow-layer features instead of the final-layer features to compute the loss. 
    To mitigate the errors caused by the discrepancies between shallow and deep-layer features, our token pruning is implemented in a gradual manner. Only a small number of tokens are pruned at the shallow layer for high fault tolerance, while more tokens are pruned at the deep layer. Furthermore, this rough estimate converges increasingly to the accuracy result (using the final layer features) as the feature layer deepens. As verified by the visualization in Fig. \ref{fig:vs}.
    \item Generating hard logit $\hat{y}_t^{l}$ as pseudo label. There is no ground truth available during the inference process, where hard logits are used as a substitute. This essentially validates the current outputs, informing the model that its current outputs are correct and that token pruning should be performed in a way that preserves these outputs.
\end{itemize}

Referring to Fig. \ref{fig:overall}, $\mathcal{P}^{l}$ generally indicates the last few tokens in the sequence, the length of which is denoted as $K_{pos}$. In the prefill stage, text tokens follow vision tokens, with access to all vision tokens and preceding text tokens. Selecting end tokens enables full use of all token information. Additionally, we divide the selected tokens into case bound and prediction tokens by their functions. The last token generates the first answer token and constrains consistent answer generation, thus being the prediction token. Modifying the prefill stage code to generate more answer tokens would make this constraint more accurate, which is beyond the scope of this paper and left for future research. The other tokens do not predict answers but constrain pruning to better adapt to the current case, hence being case bound tokens.

After obtaining the loss $L^{l}$, we perform a single-layer backward pass only with respect to the input hidden states $H^{l-1}$ of the pruning layer and compute their gradients:
\begin{equation}
	G^{l-1} = \frac{\partial L^{l}}{\partial H^{l-1}}.
\label{grandscore}
\end{equation}
We then compute the $\ell_2$-norm of the gradient vector for each token and use it as the gradient saliency score of that token at layer $l$:
\begin{equation}
s_i^{l} = \left\lVert G_i^{l-1} \right\rVert_2,
\end{equation}
where, $s_i^{l}$ measures how sensitive the layer-local proxy loss $L^{l}$ is to a local perturbation of the $i$-th token. With our tail-window design, the final position provides an intent-alignment signal, while the preceding tail positions provide a case-coverage signal, yielding an objective-driven saliency that balances contextual coverage and generation direction.

Importantly, this procedure does not modify attention operators. We only run the standard forward of the pruning layer, attach the original output head, and compute gradients with respect to $H^{l-1}$ in the usual way. Hence, the method is agnostic to the attention implementation and remains unchanged under standard attention or optimized kernels such as FlashAttention, making it fully compatible with high-efficiency attention backends in practice.

\subsection{Gradient-based Visual Token Selection with NMS}
In Sec. \ref{sec3.2}, we obtain gradient saliency scores $\{s_i^{l}\}$ at pruning layer $l$. Under a given budget $K$, TRIO selects tokens that are \emph{important} yet \emph{non-redundant} from the target compression interval . Naive Top-$K$ by $s_i^{l}$ often yields locally clustered high-score tokens, reducing global coverage under a fixed budget. To mitigate this, we apply feature-space non-maximum suppression (NMS) on top of gradient ranking to balance \emph{objective relevance} and \emph{diversity}.

Specifically, for each sample, we extract the score vector $s$ within the compression interval and sort tokens in descending order to obtain the candidate sequence $order$. We then gather the corresponding token features $v_j$ from the same interval, taken from the pruning layer input which is the previous layer hidden states, and apply $\ell_2$ normalization.

\begin{equation}
u_j=\frac{v_j}{\lVert v_j\rVert_2}.
\end{equation}
We then iteratively compute an extremely sparse upper triangular  similarity matrix, the element of which is:
\begin{equation}
S_{ij}=\langle u_i,\ u_j\rangle .
\end{equation}

Given a threshold $\tau$, we regard two tokens as highly similar in feature space when $S_{ij}\ge \tau$, and treat them as redundant neighbors.

Based on gradient ranking and this adjacency relation, we adopt a \textbf{two-stage selection strategy}.
We first perform \emph{strict NMS}: candidates are scanned in \textit{order}; if a candidate is not suppressed, it is added to the selected set $\mathcal{S}$, and all its neighbors with similarity above $\tau$ are suppressed. The process stops once $|\mathcal{S}|$ reaches the budget $K$, prioritizing high-gradient tokens while preventing feature-space clustering.

If $|\mathcal{S}|<K$ after strict NMS, we apply \emph{gradient completion} by filling the remaining slots with the highest-ranked unselected candidates following \textit{order} until the budget is met. This avoids under-selection under a high similarity threshold and ensures full budget utilization. The corresponding Algorithm is presented in Algorithm\ref{alg:nms_selection}.

This design is motivated by two points: (i) gradient saliency directly reflects each token's contribution to the output objective, while naive Top-$K$ tends to concentrate on locally high-response regions; and (ii) NMS explicitly removes redundancy via feature similarity, yielding better coverage and more stable selection under the same budget. Finally, the procedure is token-type agnostic, it may also be used to filter out text tokens, a point that will be verified in future work.
\begin{algorithm}[h]
   \caption{Gradient-based Visual Token Selection with NMS}
   \label{alg:nms_selection}
\begin{algorithmic}[1]
   \STATE {\bfseries Input:} Gradient saliency scores $\mathbf{s} \in \mathbb{R}^N$, Token features $\mathbf{V} \in \mathbb{R}^{N \times d}$, Budget $k_{keep}$, Threshold $\tau$.
   \STATE {\bfseries Output:} Selected token indices $\mathcal{S}$.

   \STATE \textit{// 1. Pre-processing}
   \STATE Normalize features: $\mathbf{u}_i \leftarrow \mathbf{v}_i / \|\mathbf{v}_i\|_2, \forall i \in \{1, \dots, N\}$
   \STATE Sort candidates: $\mathcal{O} \leftarrow \text{Argsort}(\mathbf{s}, \text{descending})$
   \STATE Initialize selected set $\mathcal{S} \leftarrow \emptyset$ and suppression mask $\mathcal{M} \leftarrow \mathbf{0}^N$

   \STATE \textit{// 2. Stage I: Strict NMS}
   \FOR{$i$ in $\mathcal{O}$}
       \IF{$|\mathcal{S}| \ge k_{keep}$}
           \STATE \textbf{break}
       \ENDIF
       \IF{$\mathcal{M}[i] = 0$}
           \STATE $\mathcal{S} \leftarrow \mathcal{S} \cup \{i\}$
           \STATE \textit{// Compute similarity with all tokens}
           \STATE $\mathbf{sim} \leftarrow \mathbf{U} \cdot \mathbf{u}_i^\top$
           \STATE \textit{// Suppress neighbors with high similarity}
           \STATE $\mathcal{M} \leftarrow \mathcal{M} \lor (\mathbf{sim} \ge \tau)$
       \ENDIF
   \ENDFOR

   \STATE \textit{// 3. Stage II: Gradient Completion (Fill remaining budget)}
   \IF{$|\mathcal{S}| < k_{keep}$}
       \FOR{$i$ in $\mathcal{O}$}
           \IF{$|\mathcal{S}| \ge k_{keep}$}
               \STATE \textbf{break}
           \ENDIF
           \IF{$i \notin \mathcal{S}$}
               \STATE $\mathcal{S} \leftarrow \mathcal{S} \cup \{i\}$
           \ENDIF
       \ENDFOR
   \ENDIF

   \STATE \textbf{return} $\mathcal{S}$
\end{algorithmic}
\end{algorithm}

\subsection{Theoretical Analysis}
\subsubsection{ Gradient Saliency and Output Invariance}
Under standard assumptions: 
(1) as shown in \cite{kim2021lipschitz} and \cite{wang2022understanding}, Transformer layers and the softmax function are Lipschitz continuous; 
(2) the gradient of the layer-local proxy loss $L_l$ with respect to hidden states characterizes how token perturbations affect the current loss and the final logits; 
(3) token pruning can be viewed as setting the hidden states of removed tokens to zero, i.e., a sparse perturbation.

Let the index set of the removed tokens at pruning layer $l$ be
\begin{equation}
\mathcal{D}^l=\{i \mid i \in V_{\text{pruned}}\}.
\end{equation}
By the Lipschitz continuity of the layer mappings, the hidden-state perturbation caused by pruning propagates to the final logits. Then, combining a first-order expansion of the layer-local proxy loss with the Cauchy--Schwarz inequality, the logit perturbation can be further related to the gradient saliency of the removed tokens. As a result, the output distribution shift can be bounded in the form
\begin{equation}
\mathrm{KL}
\left(
P(Y\mid X,V)
\|
P(Y\mid X,\widetilde{V})
\right)
\le
C
\left(
\sum_{i\in\mathcal{D}^l}
s_i^l
\|h_i^{l-1}\|_2
\right)^2 ,
\label{eq:kl_bound_main}
\end{equation}
where $P(Y\mid X,V)$ and $P(Y\mid X,\widetilde{V})$ denote the model output distributions before and after pruning, respectively. Here, $Y$ is the model output, $X$ is the textual input, and $V$ and $\widetilde{V}$ represent the visual token sets before and after pruning. Moreover, $s_i^l$ is the gradient saliency used in TRIO, and $C$ is a constant determined by the model parameters and local smoothness. Since pruning does not change the model parameters, $C$ remains unchanged.

According to the above formula, pruning low-saliency tokens makes the upper bound as small as possible, thereby better controlling the distribution shift caused by pruning. This provides theoretical support for the core intuition of TRIO, i.e., retaining visual tokens that are more important to the current inference objective according to gradient saliency. The full assumptions, derivation, and proof details are provided in Appendix~\ref{app:kl_bound}.

\subsubsection{Complexity Analysis}
Following FastV~\cite{chen2024FastV}, we count only the dominant matrix-multiplication costs in a Transformer block:
\begin{equation}
f(n)=4nd^{2}+2n^{2}d+2ndm,
\end{equation}
where $n$ is the sequence length, $d$ is the hidden dimension, and $m$ is the FFN intermediate dimension.

For a 32-layer LLaVA decoder, we prune at $[1,10,15]$, yielding [$V_0,V_1,V_2,V_3$] (effective from the next layer).
The total FLOPs are
\begin{equation}
F_{\mathrm{total}}=F_{\mathrm{inf}}+F_{\mathrm{ov}},
\end{equation}
where $F_{\mathrm{inf}}$ is the post-pruning inference cost and $F_{\mathrm{ov}}$ is the overhead(local gradients + feature-space NMS). Here $\gamma$ is the compute factor of one local backward pass relative to a forward pass:
\begin{equation}
F_{\mathrm{inf}}=f(V_0)+9f(V_1)+5f(V_2)+17f(V_3),
\end{equation}
\begin{equation}
F_{\mathrm{ov}}=\gamma\big(f(V_0)+f(V_1)+f(V_2)\big).
\end{equation}
The theoretical reduction ratio is
\begin{equation}
\mathrm{Saved}=1-\frac{F_{\mathrm{total}}}{32f(V_0)}.
\end{equation}
A detailed proof is provided in Appendix~\ref{Analysis}.

\begin{table*}[t]
\centering
\caption{\textbf{Performance comparison of different token reduction methods on LLaVA-1.5-7B under multiple token budgets.} Both \emph{w/o VE} and \emph{w/ VE} versions achieve the best results}
\label{tab:llava15_main}

\scriptsize
\setlength{\tabcolsep}{3pt}
\renewcommand{\arraystretch}{1.06}

\resizebox{\textwidth}{!}{%
\begin{tabular}{l|l|l|cccccccc|c}
\toprule
\textbf{Type} & \textbf{Methods} & \textbf{Venue} & \textbf{GQA} & \textbf{MMB} & \textbf{MMB-cn} & \textbf{MME} & \textbf{POPE} & \textbf{SQA} & \textbf{VQAv2} & \textbf{TextVQA} & \textbf{Average} \\
\midrule

\rowcolor{HeaderGray}
& \textbf{LLaVA-1.5-7B} & & \multicolumn{8}{c|}{\textbf{Upper Bound, 576 Tokens}} & \\
Baseline & Vanilla & --
& 61.9 & 64.7 & 58.1 & 1862 & 85.9 & 69.5 & 78.5 & 58.2 & \textbf{100\%} \\
\midrule

\rowcolor{HeaderGray}
& \textbf{LLaVA-1.5-7B} & & \multicolumn{8}{c|}{\textbf{Retain 192 Tokens (33.3\%)}} & \\
\multirow{5}*{\emph{w/o VE}} & FastV & ECCV'24
& 52.7 & 61.2 & 57.0 & 1612 & 64.8 & 67.3 & 67.1 & 52.5 & 89.0\% \\
 & PDrop & CVPR'25
& 57.1 & 63.2 & 56.8 & 1766 & 82.3 & 68.8 & 75.1 & 56.1 & 96.2\% \\
 & SparseVLM & ICML'25
& 59.5 & 64.1 & 53.7 & 1787 & 85.3 & 68.7 & 75.6 & 57.8 & 97.2\% \\
 & DART & EMNLP'25
& 60.0 & 63.6 & 57.0 & 1856 & 82.8 & 69.8 & 76.7 & 57.4 & 98.3\% \\
\rowcolor{OursBlue}
 & TRIO (Ours) & Ours
& 61.0 & 64.4 & 57.6 & 1789 & 86.5 & 69.0 & 77.7 & 57.2 & \textbf{98.8\%} \\
\midrule
\multirow{4}*{\emph{w/ VE}} & VisionZip & CVPR'25
& 59.3 & 63.0 & 57.3 & 1782 & 85.3 & 68.9 & 76.8 & 57.3 & 97.8\% \\
& HoloV & NeurIPS'25
& 59.0 & 65.4 & 58.0 & 1820 & 85.6 & 69.8 & 76.7 & 57.4 & 98.7\% \\
 & SCOPE & NeurIPS'25
& 60.1 & 63.6 & 56.8 & 1804 & 86.4 & 68.8 & 77.2 & 57.7 & 98.3\% \\
\rowcolor{OursBlue}
 & TRIO (Ours) & Ours
& 61.1 & 64.2 & 57.9 & 1808 & 86.4 & 68.2 & 77.9 & 57.4 & \textbf{98.9\%} \\
\midrule

\rowcolor{HeaderGray}
& \textbf{LLaVA-1.5-7B} & & \multicolumn{8}{c|}{\textbf{Retain 128 Tokens (22.2\%)}} & \\
\multirow{5}*{\emph{w/o VE}} & FastV & ECCV'24
& 49.6 & 56.1 & 56.4 & 1490 & 59.6 & 60.2 & 61.8 & 50.6 & 83.2\% \\
 & PDrop & CVPR'25
& 56.0 & 61.1 & 56.6 & 1644 & 82.3 & 68.3 & 72.9 & 55.1 & 94.0\% \\
 & SparseVLM & ICML'25
& 58.4 & 64.5 & 51.1 & 1746 & 85.0 & 68.6 & 73.8 & 56.7 & 95.6\% \\
 & DART & EMNLP'25
& 58.7 & 63.2 & 57.5 & 1840 & 80.1 & 69.1 & 75.9 & 56.4 & 97.0\% \\
\rowcolor{OursBlue}
 & TRIO (Ours) & Ours
& 60.0 & 62.9 & 57.1 & 1807 & 86.7 & 68.5 & 76.5 & 57.2 & \textbf{98.1\%} \\
\midrule
\multirow{4}*{\emph{w/ VE}} & VisionZip & CVPR'25
& 57.6 & 62.0 & 56.7 & 1761.7 & 83.2 & 68.9 & 75.6 & 56.8 & 96.4\% \\
 & HoloV & NeurIPS'25
& 57.7 & 63.9 & 56.5 & 1802 & 84.0 & 69.8 & 75.5 & 56.8 & 97.2\% \\
 & SCOPE & NeurIPS'25
& 59.7 & 62.5 & 56.9 & 1776 & 86.1 & 68.4 & 76.5 & 57.2 & 97.5\% \\
\rowcolor{OursBlue}
 & TRIO (Ours) & Ours
& 60.0 & 62.9 & 56.7 & 1799 & 86.4 & 69.2 & 77.1 & 57.0 & \textbf{98.1\%} \\
\midrule

\rowcolor{HeaderGray}
& \textbf{LLaVA-1.5-7B} & & \multicolumn{8}{c|}{\textbf{Retain 64 Tokens (11.1\%)}} & \\
\multirow{5}*{\emph{w/o VE}} & FastV & ECCV'24
& 46.1 & 48.0 & 52.7 & 1256 & 48.0 & 51.1 & 55.0 & 47.8 & 74.0\% \\
  & PDrop & CVPR'25
& 41.9 & 33.3 & 50.5 & 1092 & 55.9 & 68.6 & 69.2 & 45.9 & 74.4\% \\
& SparseVLM & ICML'25
& 53.8 & 60.1 & 52.7 & 1589 & 77.5 & 69.8 & 68.2 & 53.4 & 90.6\% \\
 & DART & EMNLP'25
& 55.9 & 60.6 & 53.2 & 1765 & 73.9 & 69.8 & 72.4 & 54.4 & 92.8\% \\
\rowcolor{OursBlue}
 & TRIO (Ours) & Ours
& 58.0 & 61.6 & 53.7 & 1681 & 84.3 & 68.5 & 74.8 & 54.9 & \textbf{94.7\%} \\
\midrule
\multirow{4}*{\emph{w/ VE}} & VisionZip & CVPR'25
& 55.1 & 60.1 & 55.4 & 1690 & 77.0 & 69.0 & 72.4 & 55.5 & 93.0\% \\
& HoloV & NeurIPS'25
& 55.3 & 63.3 & 55.1 & 1715 & 80.3 & 69.5 & 72.8 & 55.4 & 94.4\% \\
 & SCOPE & NeurIPS'25
& 58.3 & 61.7 & 54.4 & 1698 & 83.9 & 68.6 & 75.3 & 56.6 & 95.4\% \\
\rowcolor{OursBlue}
 & TRIO (Ours) & Ours
& 58.3 & 61.6 & 56.5 & 1744 & 86.4 & 68.6 & 75.9 & 56.2 & \textbf{96.6\%} \\
\bottomrule
\end{tabular}
} 
\end{table*}

\begin{table*}[t]
\centering
\caption{\textbf{Performance comparison of different token reduction methods on LLaVA-NeXT-7B under multiple token budgets.} Blue-highlighted rows denote our method.}
\label{tab:llava16_main}
\small
\setlength{\tabcolsep}{4pt}
\renewcommand{\arraystretch}{1.08}

\begin{tabular}{l|l|cccccccc|c}
\toprule
\textbf{Method} & \textbf{Venue} & \textbf{GQA} & \textbf{MMBench} & \textbf{MMBench-cn} & \textbf{MME} & \textbf{POPE} & \textbf{SQA} & \textbf{VQAv2} & \textbf{TextVQA} & \textbf{Avg} \\
\midrule

\rowcolor{HeaderGray}
\textbf{LLaVA-NeXT-7B}&  & \multicolumn{8}{c|}{\textbf{Upper Bound, 2880 Tokens}} & \\
Vanilla & --
& 64.2 & 67.4 & 60.6 & 1851 & 86.5 & 70.1 & 81.8 & 61.4 & 100\% \\
\midrule

\rowcolor{HeaderGray}
\textbf{LLaVA-NeXT-7B}& & \multicolumn{8}{c|}{\textbf{Retain 320 Tokens}} & \\
FastV & ECCV'24
& 55.9 & 61.6 & 51.9 & 1661 & 71.7 & 62.8 & 71.9 & 55.7 & 88.1\% \\
SparseVLM & ICML'25
& 56.1 & 60.6 & 54.5 & 1533 & 82.4 & 66.1 & 71.5 & 58.4 & 90.3\% \\
VisionZip & CVPR'25
& 59.3 & 63.1 & 55.6 & 1702 & 82.1 & 67.3 & 76.2 & 58.9 & 93.7\% \\
CDPruner & NeurIPS'25
& 61.6 & 65.5 & 55.7 & 1453 & 87.2 & 67.8 & 78.4 & 57.4 & 93.8\% \\
DART & EMNLP'25
& 61.7 & 65.3 & 58.2 & 1710 & 84.1 & 68.4 & 79.1 & 58.7 & 96.1\% \\
HoloV & NeurIPS'25
& 61.7 & 65.3 & 57.5 & 1738 & 83.9 & 68.9 & 79.5 & 58.7 & 96.2\% \\
\rowcolor{OursBlue}
TRIO (Ours) & Ours
& 61.8 & 66.2 & 59.5 & 1795 & 84.5 & 68.9 & 79.3 & 58.3 & \textbf{97.2\%} \\
\bottomrule
\end{tabular}
\end{table*}
\section{Experiments and Analysis}\label{Experiments}
\subsection{Experimental Setup}
\subsubsection{Models and Baselines}
We evaluate TRIO on three LVLMs with diverse architectures: LLaVA-1.5 \cite{liu2023llava}, LLaVA-NeXT \cite{liu2024llavanext}, and Qwen-2.5-VL \cite{bai2025qwen2}.
We compare against eight representative visual-token reduction baselines, including FastV \cite{chen2024FastV}, SparseVLM \cite{zhang2025sparsevlm}, PyramidDrop \cite{xing2024pyramiddrop}, VisionZip \cite{yang2025visionzip}, DART \cite{wen2025dart}, HoloV \cite{holov}, SCOPE \cite{deng2025scope}, and CDPruner \cite{zhang2025cdpruner}.

\subsubsection{Datasets}
We report image-based results on eight standard benchmarks: GQA \cite{hudson2019gqa}, MMBench, MMBench-CN \cite{liu2024mmbench}, MME \cite{fu2025mme}, POPE \cite{li2023pope}, SQA \cite{lu2022sqa}, VQAV2 \cite{goyal2017vqav2}, and TextVQA \cite{singh2019textvqa}.

\subsubsection{Implementation Details}
We follow the default inference settings in the official codebases.
Pruning layers are set to $[1,10,15]$ for LLaVA models and $[1,8,14]$ for Qwen models, with $K_{pos}=4$ and $\tau=0.8$.
Ablations are provided in Section~\ref{Ablation Study and Analysis}.

\subsection{Main Results}
\subsubsection{Results on LLaVA-1.5}
We first apply TRIO to LLaVA-1.5, which is widely adopted for evaluating visual token pruning strategies. As shown in Table~\ref{tab:llava15_main}, TRIO achieves state-of-the-art performance across all budgets in \textbf{both} \emph{w/ VE} and \emph{w/o VE} settings.
In the \emph{$w/o VE$} track, it consistently outperforms competitors, retaining 98.8\% and 94.7\% average performance at 192 and 64 tokens respectively, surpassing DART (98.3\% and 92.8\%).
Furthermore, the \emph{w/ VE} setting elevates the SOTA to a higher level: it reaches 98.9\% retention at 192 tokens and maintains 96.6\% under the extreme 64-token regime, significantly outperforming the previous best Encoder-Involved method. 

Notably, before the proposal of TRIO, \emph{w/o VE} methods generally exhibited lower accuracy than \emph{w/ VE} ones. Our w/o VE variant, by contrast, not only outperforms strong w/o VE baselines such as DART but also surpasses some existing w/ VE methods.
Our \emph{w/ VE} variant adopts a two-stage hybrid scheme: SCOPE-style pre-filtering \cite{deng2025scope} before token injection, followed by objective-driven refinement with TRIO at layer 16.
This combines early efficiency with stronger deep-layer objective alignment, yielding better results than pre-filtering alone under the same budget.

\begin{table*}[t]
\centering
\caption{\textbf{Performance comparison of different token reduction methods on Qwen2.5-VL-7B under multiple token budgets.} Blue-highlighted rows denote our method.}
\label{tab:qwen2.5}
\small
\setlength{\tabcolsep}{4pt}
\renewcommand{\arraystretch}{1.08}

\begin{tabular}{l|l|ccccccc|c}
\toprule
\textbf{Method} & \textbf{Venue} & \textbf{MMBench} & \textbf{MME} & \textbf{POPE} & \textbf{SQA} & \textbf{MMBench-cn} & \textbf{GQA} & \textbf{TextVQA} & \textbf{Avg} \\
\midrule

\rowcolor{HeaderGray}
\textbf{Qwen-2.5-VL}& & \multicolumn{7}{c|}{\textbf{Upper Bound, 1296 Tokens}} & \\
Vanilla & --
& 83.7 & 2299 & 87 & 88.65 & 81.8 & 61 & 77.7 & 100\% \\
\midrule

\rowcolor{HeaderGray}
\textbf{Qwen-2.5-VL}& & \multicolumn{7}{c|}{\textbf{Retain 33.30\% Tokens}} & \\
DART & EMNLP'25
& 80.9 & 2316 & 82.8 & 84.88 & 79.0 & 57.67 & 70.4 & 95.7\% \\
\rowcolor{OursBlue}
TRIO (Ours) & Ours
& 82.1 & 2317 & 85.9 & 88.7 & 80.5 & 60.1 & 75.5 & \textbf{98.8\%} \\
\midrule

\rowcolor{HeaderGray}
\textbf{Qwen-2.5-VL}& & \multicolumn{7}{c|}{\textbf{Retain 22.2\% Tokens}} & \\
DART & EMNLP'25
& 78.9 & 2227 & 80.4 & 84.2 & 76.5 & 55.76 & 67 & 92.8\% \\
\rowcolor{OursBlue}
TRIO (Ours) & Ours
& 80.9 & 2284 & 85.4 & 88 & 78.1 & 58.8 & 74.4 & \textbf{97.3\%} \\
\midrule

\rowcolor{HeaderGray}
\textbf{Qwen-2.5-VL}& & \multicolumn{7}{c|}{\textbf{Retain 11.1\% Tokens}} & \\
DART & EMNLP'25
& 73.3 & 1971 & 73.3 & 82 & 71.5 & 50.3 & 57.3 & 84.8\% \\
\rowcolor{OursBlue}
TRIO (Ours) & Ours
& 77.6 & 2150 & 77.5 & 84.8 & 75.5 & 53.5 & 70.6 & \textbf{91.7\%} \\
\bottomrule
\end{tabular}
\end{table*}

\subsubsection{Results on LLaVA-NeXT}
To assess scalability on high-resolution architectures with denser visual tokens, we simulate a resource-constrained setting by retaining 320 visual tokens on average per layer.
As shown in Table~\ref{tab:llava16_main}, TRIO remains robust under this aggressive compression, achieving 97.2\% SOTA average retention and surpassing HoloV (96.2\%) and DART (96.1\%), while substantially alleviating the severe degradation of FastV (88.1\%) and SparseVLM (90.3\%) .

\subsubsection{Results on Qwen-2.5-VL}
Following the CDPruner protocol~\cite{zhang2025cdpruner}, we fix the input resolution to $1008\times1008$ (1,296 visual tokens per image).
As shown in Table~\ref{tab:qwen2.5}, TRIO consistently outperforms DART across all pruning ratios, and the advantage becomes more pronounced at higher compression levels:
at the aggressive 11.1\% retention rate, TRIO maintains 91.7\% average performance retention, exceeding DART (84.8\%) by 6.9\% .
Overall, these results confirm that our inference-objective perspective robustly preserves discriminative semantics across diverse VLM backbones and token-density regimes.

\subsection{Ablation Study and Analysis}\label{Ablation Study and Analysis}

\begin{figure}
    \centering
    \includegraphics[width=\linewidth]{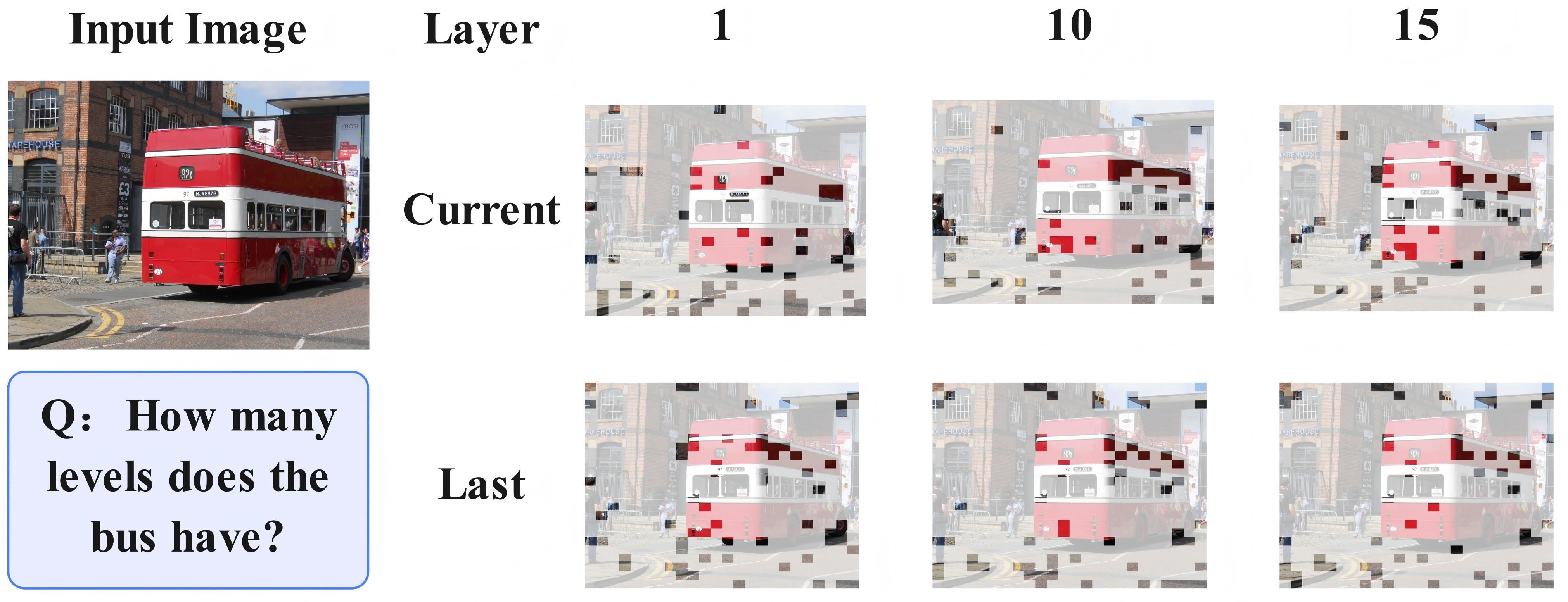}
    \caption{Visualization of selected important visual tokens (64 retained) using different gradient signals. Token selection is performed at Layers $[1,10,15]$: the top row (Current) ranks image tokens by gradients from the current-layer local proxy loss, while the bottom row (Last) uses gradients propagated from the final-layer output loss. Highlighted patches indicate the retained tokens.}
    \label{fig:vs}
\end{figure}

\subsubsection{Rationale of Gradient-Based Token Selection}
Motivated by findings~\cite{jain2019attention} that attention weights can be weakly correlated with feature importance and can be perturbed with negligible output change, we adopt \emph{gradient saliency} to measure each token's influence via output sensitivity. Fig.~\ref{fig:vs} visualizes the selected \textbf{64} visual tokens at pruning Layers~\textbf{$[1,10,15]$}, comparing two importance signals: \textbf{Current} (layer-local gradients induced by a proxy loss at the current layer) and \textbf{Last} (global gradients backpropagated from the final output loss). As illustrated in Fig.\ref{fig:vs}, two key phenomena can be observed:
\begin{itemize}
    \item The guiding signals generated by the current layer exhibit a high degree of consistency with those from the last layer. Furthermore, in practice, given that Transformer blocks possess a global receptive field, tokens that are spatially adjacent exhibit high similarity. Thus, the selected tokens do not require strict alignment; approximate effects can be achieved as long as their covered regions are roughly similar.

    \item With increasing depth, the consistency of the guiding signals from the current layer and the last layer exhibits a gradual increase. This is quite intuitive: as the depth increases, the level of feature abstraction rises, and the features of the current layer become increasingly close to those of the last layer. This characteristic aligns perfectly with our design. At shallow layers, consistency exists but remains low, so we only remove a small number of vision tokens, resulting in a high fault tolerance. At deep layers, since the guiding signals become more accurate, we can eliminate more vision tokens.
\end{itemize}

\begin{figure}
    \centering
    \includegraphics[width=\linewidth]{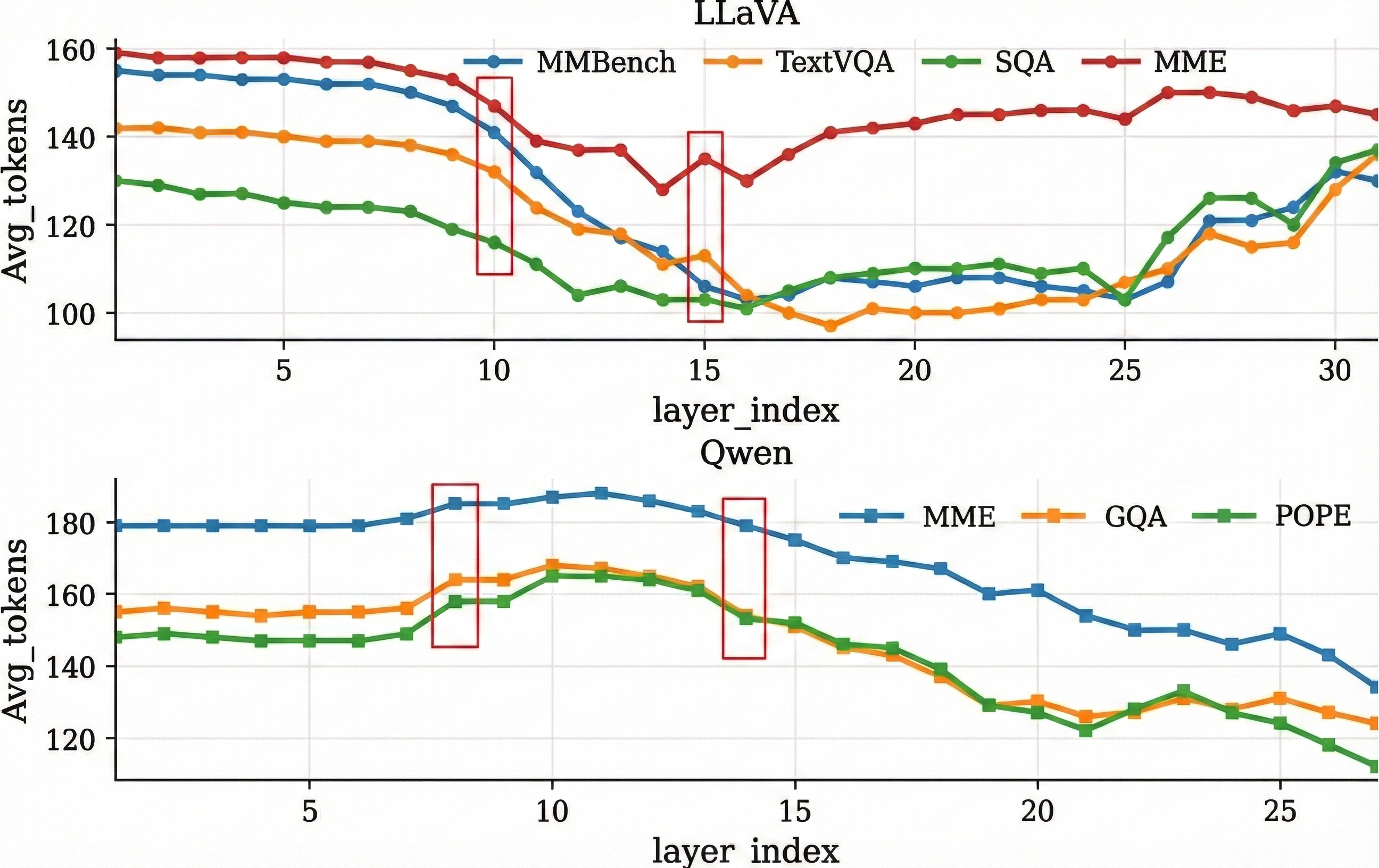}
\caption{\textbf{Layer-wise gradient saliency statistics.} For each benchmark and layer, we report the average number of tokens per question whose gradient scores exceed the layer-wise mean. Red boxes highlight layers exhibiting notable distribution shifts for LLaVA and Qwen.}
\label{fig:avg_token}
\end{figure}

\subsubsection{Hyperparameter Settings}

\paragraph{Pruning-layer selection strategy} Guided by the layer-wise gradient-saliency curves in Fig. \ref{fig:avg_token}, which illustrate the average number of tokens per layer whose gradient scores exceed the mean, we prune at Layers $[1,10,15]$ for LLaVA and Layers $[1,8,14]$ for Qwen.

This selection is primarily based on two principles:
\begin{itemize}
    \item Pruning is performed as early as possible in the network, as this most effectively reduces computational overhead and achieves acceleration. Therefore, all selected pruning layers lie in the first half of the network.
    \item We select layers with drastic gradient changes, as such layers exhibit sharp variations in token information, and this is exactly when tokens need to be selected carefully.
\end{itemize}

Importantly, TRIO does not rely on one exact set of layer indices. The key is to follow a stable pruning pattern: performing an early pruning step to reduce the sequence length as soon as possible, and placing the subsequent pruning layers around the middle stages where token representations become more semantically stable. To further verify this stability, we report the sensitivity results for nearby pruning-layer choices in Appendix~\ref{app:layer_sensitivity}. The results indicate that shifting the pruning layers to adjacent indices only causes minor performance variations, suggesting that our method is not tied to a precise layer configuration. Although the optimal indices may vary slightly across different backbones, the overall pattern remains consistent; therefore, only light adjustment around neighboring layers is typically needed, rather than a full hyperparameter search.

\paragraph{$K_{\text{pos}}$ selection}
As shown in Table~\ref{tab:kpos_ablation_avg}, the choice of $K_{\text{pos}}$ substantially affects the stability and coverage of the proxy loss. Using only the last position ($K_{\text{pos}}{=}1$) yields a sparse supervision signal and results in a much lower average accuracy ($92.1\%$). Once a few preceding positions are included ($K_{\text{pos}}{=}2\sim5$), the proxy loss both aligns with the immediate output objective and better covers the current instance, leading to a large performance jump and a stable plateau. In particular, $K_{\text{pos}}{=}4$ achieves the best average accuracy ($94.9\%$), offering a more reasonable trade-off between objective alignment and contextual robustness. Further increasing $K_{\text{pos}}$ provides diminishing returns and can slightly degrade performance, suggesting that too many positions may introduce noisier and less focused gradient signals.

\begin{table}
\centering
\caption{\textbf{Ablation on $K_{\text{pos}}$ (Avg scores on different benchmarks).} All settings use the same pruning configuration (LLaVA-v1.5-7B, tokens=64); higher is better.}
\label{tab:kpos_ablation_avg}
\small
\setlength{\tabcolsep}{2pt}
\renewcommand{\arraystretch}{1.15}

\begin{tabular*}{\columnwidth}{@{\extracolsep{\fill}}c|cccccccc}
\toprule
$K_{\text{pos}}$ & 1 & 2 & 3 & 4 & 5 & 6 & 7 & 8 \\
\midrule
Avg & \textit{92.1\%} & \textit{94.6\%} & \textit{94.6\%} & \textit{94.9\%} & \textit{94.9\%} & \textit{94.8\%} & \textit{94.7\%} & \textit{94.8\%} \\
\bottomrule
\end{tabular*}
\end{table}

\begin{table}
\centering
\caption{\textbf{Ablation on NMS and $\tau$}  \textbf{Avg scores on different benchmarks).} All settings use the same pruning configuration(llava-v1.5-7B,tokens=64) (higher is better) .}
\label{tab:nms_tau_avg}
\small
\setlength{\tabcolsep}{2pt}
\renewcommand{\arraystretch}{1.08}
\begin{tabular}{c|ccccccc}
\toprule
$\boldsymbol{\tau}$ & 0.70 & 0.75 & 0.80 & 0.85 & 0.90 & 0.95&no nms \\
\midrule
\textbf{Avg} & {94.6\%} & {94.7\%} & {94.8\%} & {94.4\%} & {94.2\%} & {93.6\%} &{92.1\%} \\
\bottomrule
\end{tabular}
\end{table}

\paragraph{NMS and threshold choice}
As shown in Table~\ref{tab:nms_tau_avg}, applying feature-space cosine-similarity NMS consistently improves performance over the Top-$K$ baseline without NMS. As the threshold $\tau$ increases, suppression becomes weaker: overly small $\tau$ may over-prune and discard useful evidence, while overly large $\tau$ fails to remove redundant, highly similar tokens and degenerates toward Top-$K$ behavior. Empirically, $\tau=0.8$ yields the best average accuracy ($94.8\%$), suggesting a better trade-off between reducing redundancy for broader coverage and preserving critical information.

\begin{table*}[t]
\centering
\small
\setlength{\tabcolsep}{5pt}
\renewcommand{\arraystretch}{1.12}
\caption{
\textbf{Summary of performance and efficiency comparisons.}
Panel (a) reports the average performance across all benchmarks under matched visual-token budgets.
Panel (b) reports the efficiency on the POPE benchmark with LLaVA-NeXT-7B at 11.1\% visual-token retention.
Panel (c) compares the overall runtime speedup on LLaVA-NeXT-7B under the same retention setting.
}
\label{tab:summary_efficiency}
\begin{tabular}{llccc}
\toprule
\textbf{Model} & \textbf{Tokens} & \textbf{Ours} & \textbf{Grid} & \textbf{Random} \\
\midrule
\multicolumn{5}{c}{\textit{(a) Average performance across all benchmarks under matched visual-token budgets}} \\
\midrule
LLaVA-1.5-7B   & 192 & \textbf{98.8} & 96.3 & 95.7 \\
LLaVA-1.5-7B   & 128 & \textbf{98.2} & 94.9 & 93.4 \\
LLaVA-1.5-7B   & 64  & \textbf{94.7} & 88.9 & 87.0 \\
LLaVA-NeXT-7B  & 320 & \textbf{96.5} & 91.2 & 90.4 \\
\midrule
\multicolumn{5}{c}{\textit{(b) Efficiency on POPE with LLaVA-NeXT-7B at 11.1\% visual-token retention}} \\
\midrule
\textbf{Method} & \textbf{FLOPs (T)} & \textbf{Total (s)} & \textbf{Prefill (s)} & \textbf{KV (MB)} \\
Baseline & 16.67  & 5921 & 4934 & 1156 \\
Ours     & 2.68   & 2764 & 1791 & 191  \\
Grid     & 1.9038 & 2256 & 1297 & 191  \\
Random   & 1.9037 & 2261 & 1302 & 191  \\
\midrule
\multicolumn{5}{c}{\textit{(c) Overall runtime speedup on LLaVA-NeXT-7B at 11.1\% visual-token retention}} \\
\midrule
\textbf{Method} & \multicolumn{4}{c}{\textbf{Overall Runtime Speedup Ratio}} \\
Ours      & \multicolumn{4}{c}{\textbf{2.14$\times$}} \\
SparseVLM & \multicolumn{4}{c}{1.56$\times$} \\
DART      & \multicolumn{4}{c}{1.99$\times$} \\
\bottomrule
\end{tabular}
\end{table*}

\paragraph{Performance--Efficiency Trade-off}

Table~\ref{tab:summary_efficiency}(a) first compares TRIO with Random Sampling and Uniform Grid Sampling under the same visual-token retention ratios. TRIO consistently achieves better performance preservation across all token budgets. The advantage becomes more evident under aggressive compression. For example, when retaining only 64 visual tokens on LLaVA-1.5-7B, Grid and Random drop to 88.9 and 87.0 in average performance, respectively, while TRIO still maintains 94.7. This indicates that simple sampling strategies, despite their negligible selection overhead, cannot reliably preserve task-critical visual tokens.

Table~\ref{tab:summary_efficiency}(b) further reports the efficiency results on POPE with LLaVA-NeXT-7B at 11.1\% visual-token retention. It is worth noting that Grid and Random can be regarded as purely idealized sampling baselines with almost no token-importance estimation process, and therefore their runtime is closer to a low-overhead lower bound. In contrast, TRIO computes gradient saliency through a local backward pass, leading to a slightly higher total runtime of 2764s, compared with 2256s for Grid and 2261s for Random. However, this extra overhead accounts for less than 9\% of the original full-token baseline runtime, while bringing substantially better performance preservation. In other words, TRIO does not aim to be faster than these purely idealized sampling baselines; instead, it achieves more reliable token selection with acceptable extra overhead, while still reducing the total runtime from 5921s to 2764s compared with the full-token baseline.

More importantly, although TRIO introduces a local backward overhead, it still achieves faster overall inference than existing state-of-the-art token compression methods. As shown in Table~\ref{tab:summary_efficiency}(c), under the same LLaVA-NeXT-7B setting with 11.1\% visual-token retention, TRIO achieves a $2.14\times$ overall runtime speedup, outperforming SparseVLM with $1.56\times$ and DART with $1.99\times$. This demonstrates that the additional gradient-computation cost can be effectively compensated by the computational savings from reducing the sequence length in subsequent layers. Moreover, TRIO does not modify the attention operator and is naturally compatible with FlashAttention, leaving further room for practical deployment optimization. Overall, TRIO is slightly slower than purely idealized sampling baselines but achieves much better performance preservation; compared with existing SOTA methods, it still delivers faster overall inference even with the additional backward overhead.Additional efficiency analysis is provided in Appendix~\ref{app:efficiency_analysis}.

\section{Conclusion}\label{Conclusion}

This paper presents TRIO, a training-free acceleration method for VLMs. Unlike existing methods that mainly rely on attention maps or feature-similarity metrics to estimate visual token importance, TRIO, to the best of our knowledge, is the first to leverage gradient saliency from the inference objective to guide visual token selection. Specifically, we formulate token compression as an output-invariance preservation problem and design a layer-local proxy loss to approximate the current inference objective, thereby deriving token-level gradient signals that are more directly related to the model output. Based on this gradient saliency, TRIO reorders and selects visual tokens, reducing redundant tokens while preserving task-critical visual information. Since our method does not modify the attention operator or require explicit attention-map extraction, it is naturally compatible with FlashAttention and can serve as a plug-and-play training-free acceleration module for different VLM backbones.

\FloatBarrier
\section*{Acknowledgments}
This work was supported in part by the National Natural Science Foundation
of China under Grant 62401471, in part by 2024 Gusu Innovation and
Entrepreneurship Leading Talents Program under Grant ZXL2024333.

\bibliographystyle{IEEEtran}
\bibliography{reference}

\clearpage
\appendices
\addcontentsline{toc}{section}{Appendix} 
\setcounter{section}{0}  

\section{Visual Comparison and Failure Mode Analysis of Four Token Selection Strategies}
\label{sec:vis_case_study_en}

\begin{figure*}[h]
    \centering
    \includegraphics[width=0.9\linewidth]{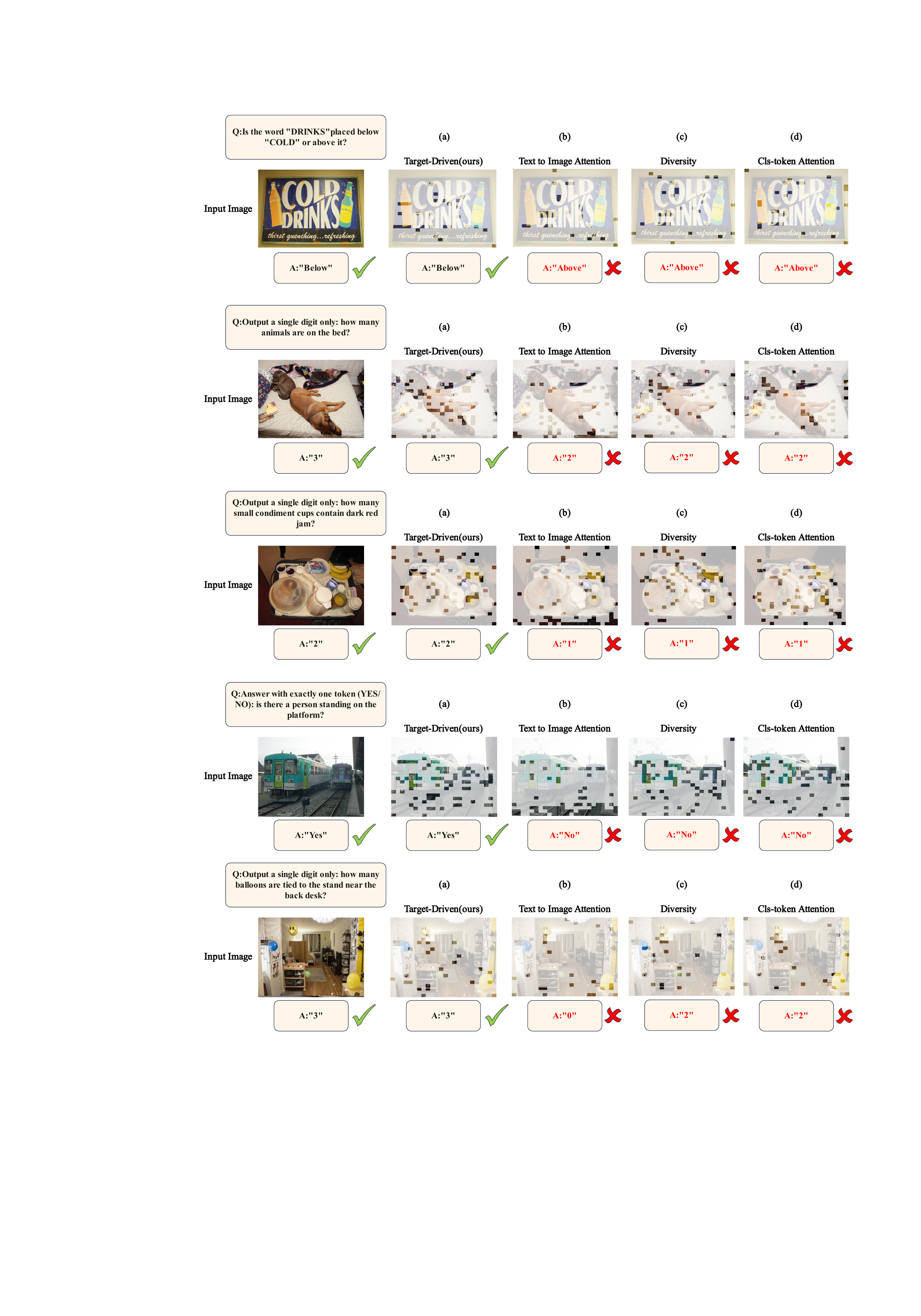}
    \caption{Comparison of different token selection strategies . (a) Ours; (b) Similirty with cls token and text token based; (c) cls token similarity based and enhance diversity; (d) Cls token similarity based.}
    \label{fig:four-methods2}
\end{figure*}

\paragraph{Qualitative Analysis of Retained Evidence}
To reveal how different visual token reduction strategies affect the retained evidence, we compare four representative methods under the same input and the same token budget, as shown in Fig.~\ref{fig:four-methods2}. The four methods include objective driven selection, text to image attention based selection, diversity oriented selection, and CLS attention based selection. Each question corresponds to one row of results, where colored grids indicate the retained visual token locations, together with the predicted answers and correctness marks.

Overall, the objective driven method consistently allocates the limited budget to regions that are directly relevant to the question, such as key fragments of scene text, local areas required for spatial relation reasoning, and dense regions of objects for counting. As a result, it remains more stable on fine grained tasks that rely on precise evidence, including text reading and relative position judgment, object counting, and existence verification. In contrast, attention based heuristics can be biased toward globally salient regions or language priors, which may lead to insufficient coverage near the decisive evidence and thus systematic errors. The diversity oriented strategy provides broader spatial coverage, but it dilutes the sampling density over crucial regions under a fixed budget, making it prone to misses on fine grained recognition and counting. The CLS attention strategy behaves more like a global summary and is less targeted to local discriminative cues, which also causes failures on samples requiring high resolution evidence.

These cases suggest that visual token selection should consider not only relevance but also evidence usability and discriminability under a strict budget. By aligning token selection with the current output objective, the objective driven mechanism preserves more useful evidence at the same budget, leading to more reliable answers.

\paragraph{Failure Case Analysis}
Although TRIO can better preserve task-relevant evidence in most cases, it also has certain failure cases. Since TRIO relies on the current prediction direction, i.e., the pseudo label constructed from the current model output, objective-driven selection may reinforce an unreliable hypothesis when the current prediction is incorrect or low-confidence. In such cases, similarity- or coverage-based methods may sometimes be more robust, as they preserve a broader set of visual regions.

This limitation mainly appears in two scenarios. First, for shallow-layer or hard samples with low-confidence predictions, the intermediate representation may not yet be semantically stable. If token selection is performed too early according to the current hypothesis, the model may discard visual evidence that could be useful for later correction. Second, for tasks with multi-token answers or ambiguity resolved only in later decoding steps, using only a few tail positions during prefilling to construct the proxy loss may be insufficient to capture the complete answer semantics. Incorporating more answer-side tokens could provide a more accurate objective constraint.

Therefore, the effectiveness of TRIO depends on the reliability of the current prediction direction. When the prediction confidence is low, the answer is long, or decisive evidence emerges only during later decoding steps, relying solely on the current proxy objective may be limited. Future work could further improve robustness by incorporating confidence-aware selection, longer answer-side constraints, or recoverable token selection mechanisms.

\section{Proof of the Output Distribution Shift Bound}
\label{app:kl_bound}

In this section, we provide the proof of the output distribution shift bound under visual token pruning. Let $V$ and $\widetilde{V}$ denote the visual token sets before and after pruning, respectively. The corresponding model output distributions are denoted as
\begin{equation}
P(Y|X,V),
\qquad
P(Y|X,\widetilde{V}) .
\end{equation}

Let $\mathcal{D}^l$ be the index set of visual tokens removed at pruning layer $l$:
\begin{equation}
\mathcal{D}^l
=
\{i \mid v_i \text{ is removed at layer } l\}.
\end{equation}

Pruning can be viewed as setting the hidden states of the removed tokens to zero. Therefore, for each $i\in\mathcal{D}^l$, we have
\begin{equation}
\widetilde{h}_i^{l-1}=0,
\qquad
\Delta h_i^{l-1}
=
\widetilde{h}_i^{l-1}
-
h_i^{l-1}
=
-h_i^{l-1}.
\end{equation}

Thus, the hidden-state perturbation introduced by pruning satisfies
\begin{equation}
\|\Delta h^{l-1}\|_2
\leq
\sum_{i\in\mathcal{D}^l}
\|h_i^{l-1}\|_2 .
\label{eq:hidden_perturb}
\end{equation}

Since Transformer layers, normalization layers, and the output head are locally Lipschitz continuous, the hidden-state perturbation is propagated to the final logits in a bounded manner. Let $Z$ and $\widetilde{Z}$ denote the logits before and after pruning. Then there exists a constant $C_1>0$ such that
\begin{equation}
\|Z-\widetilde{Z}\|_2
\leq
C_1
\|\Delta h^{l-1}\|_2 .
\label{eq:logit_lipschitz}
\end{equation}

However, Eq.~\eqref{eq:logit_lipschitz} only measures the magnitude of hidden-state perturbations and does not distinguish the importance of different tokens. To incorporate task-dependent sensitivity, TRIO defines the gradient saliency of token $i$ at layer $l$ as
\begin{equation}
s_i^l
=
\left\|
\frac{\partial L^l}
{\partial h_i^{l-1}}
\right\|_2 ,
\label{eq:saliency_def}
\end{equation}
where $L^l$ is the layer-local proxy loss.

By the first-order Taylor expansion, the change of the proxy loss caused by pruning can be approximated as
\begin{equation}
\Delta L^l
\approx
\sum_{i\in\mathcal{D}^l}
\left\langle
\frac{\partial L^l}{\partial h_i^{l-1}},
\Delta h_i^{l-1}
\right\rangle .
\end{equation}

Applying the Cauchy--Schwarz inequality gives
\begin{equation}
|\Delta L^l|
\leq
\sum_{i\in\mathcal{D}^l}
\left\|
\frac{\partial L^l}
{\partial h_i^{l-1}}
\right\|_2
\|\Delta h_i^{l-1}\|_2 .
\end{equation}

Since $\|\Delta h_i^{l-1}\|_2=\|h_i^{l-1}\|_2$ for $i\in\mathcal{D}^l$, we obtain
\begin{equation}
|\Delta L^l|
\leq
\sum_{i\in\mathcal{D}^l}
s_i^l
\|h_i^{l-1}\|_2 .
\label{eq:loss_bound}
\end{equation}

Because the layer-local proxy loss is constructed to reflect the current inference objective, its gradient characterizes how token perturbations affect the current loss and the final logits. Therefore, there exists a constant $C_2>0$ such that the logit perturbation can be bounded by the saliency-weighted perturbation:
\begin{equation}
\|Z-\widetilde{Z}\|_2
\leq
C_2
\sum_{i\in\mathcal{D}^l}
s_i^l
\|h_i^{l-1}\|_2 .
\label{eq:logit_saliency_bound}
\end{equation}

Next, since the softmax function and the induced output distribution are locally smooth with respect to logits, there exists a constant $C_3>0$ such that
\begin{equation}
\mathrm{KL}
\left(
P(Y|X,V)
\|
P(Y|X,\widetilde{V})
\right)
\leq
C_3
\|Z-\widetilde{Z}\|_2^2 .
\label{eq:kl_logit_bound}
\end{equation}

Substituting Eq.~\eqref{eq:logit_saliency_bound} into Eq.~\eqref{eq:kl_logit_bound}, we have
\begin{equation}
\mathrm{KL}
\left(
P(Y|X,V)
\|
P(Y|X,\widetilde{V})
\right)
\leq
C_3 C_2^2
\left(
\sum_{i\in\mathcal{D}^l}
s_i^l
\|h_i^{l-1}\|_2
\right)^2 .
\end{equation}

Let $C=C_3C_2^2$. The final bound is
\begin{equation}
\mathrm{KL}
\left(
P(Y|X,V)
\|
P(Y|X,\widetilde{V})
\right)
\leq
C
\left(
\sum_{i\in\mathcal{D}^l}
s_i^l
\|h_i^{l-1}\|_2
\right)^2 .
\label{eq:kl_bound_appendix}
\end{equation}

The constant $C$ is determined by the local smoothness of the fixed model. Since pruning does not modify model parameters, $C$ remains unchanged during pruning. Eq.~\eqref{eq:kl_bound_appendix} indicates that the output distribution shift is mainly controlled by the accumulated saliency cost of the removed tokens. Therefore, removing low-saliency tokens leads to a smaller upper bound, which theoretically supports the gradient-guided token selection strategy of TRIO.

\section{Detailed Experiment Settings.}\label{Detailed Experiment Settings.}
\subsection{Benchmarks and Models}

\paragraph{Benchmarks}
We evaluate TRIO on a diverse set of widely used vision--language benchmarks, covering general visual question answering, compositional reasoning, OCR-centric understanding, hallucination evaluation, and comprehensive multimodal capability assessment.

\textbf{GQA}~\cite{hudson2019gqa} is a large-scale visual question answering benchmark built upon structured scene graphs. It emphasizes object attributes, relations, and compositional reasoning, making it suitable for evaluating fine-grained visual grounding and structured visual understanding.

\textbf{TextVQA}~\cite{singh2019textvqa} focuses on visual question answering that requires reading and reasoning over scene text. Since many questions depend on text appearing in natural images, it is commonly used to evaluate OCR-centric understanding and text-aware multimodal reasoning.

\textbf{ScienceQA}~\cite{lu2022sqa} is a multimodal scientific question answering benchmark containing multiple-choice questions from diverse science domains. It evaluates whether models can combine visual information, textual context, and background knowledge for structured reasoning.

\textbf{POPE}~\cite{li2023pope} is designed to evaluate object-level hallucination in vision--language models. It formulates binary questions about object existence and measures whether a model produces answers that are faithful to the visual content.

\textbf{MME}~\cite{fu2025mme} provides a comprehensive evaluation of multimodal large language models. It covers both perception-oriented tasks, such as recognition and counting, and cognition-oriented tasks, such as commonsense and logical reasoning.

\textbf{VQAv2}~\cite{goyal2017vqav2} is a standard visual question answering benchmark where models answer natural-language questions about images. By introducing paired images with different answers for the same question, it helps reduce language priors and encourages visually grounded predictions.

\textbf{MMBench}~\cite{liu2024mmbench} evaluates multimodal understanding and reasoning through multiple-choice questions. It covers diverse abilities such as object recognition, spatial reasoning, attribute understanding, commonsense reasoning, and logical inference, enabling fair and automatic cross-model comparison.

\paragraph{Models}
We conduct experiments on three representative vision--language model backbones to evaluate both effectiveness and generalization of TRIO.

\textbf{LLaVA-1.5}~\cite{liu2023llava} follows the standard architecture of a vision encoder, a multimodal projector, and an LLM decoder. It is a widely used open-source VLM baseline and has been extensively adopted in studies on multimodal reasoning and visual token reduction.

\textbf{LLaVA-NeXT}~\cite{liu2024llavanext} is an improved model in the LLaVA family, with stronger instruction-following ability and better support for high-resolution image inputs. Since high-resolution inputs usually introduce longer visual-token sequences, LLaVA-NeXT serves as a more challenging benchmark for evaluating inference acceleration methods.

\textbf{Qwen2.5-VL}~\cite{bai2025qwen2} is a strong multimodal large language model with broad capabilities in image question answering, OCR-centric understanding, visual grounding, and multi-step reasoning. We use it to examine whether TRIO can generalize beyond the LLaVA-style architecture.

\begin{table*}[t]
\centering
\small
\setlength{\tabcolsep}{4pt}
\renewcommand{\arraystretch}{1.12}
\caption{
\textbf{Overall efficiency and local timing results on LLaVA-NeXT-7B.}
We report FLOPs, total runtime, prefill time, KV-cache memory, and the local timing breakdown introduced by TRIO.
}
\label{tab:overall_efficiency_local_timing}
\resizebox{\textwidth}{!}{
\begin{tabular}{lccccccc}
\toprule
\textbf{Method} 
& \textbf{Avg FLOPs (T)}
& \textbf{Total Time (s)}
& \textbf{Prefill Time (s)}
& \textbf{KV Cache (MB)}
& \textbf{Forward Total (ms)}
& \textbf{Proxy Loss Total (ms)}
& \textbf{Backward Total (ms)} \\
\midrule
LLaVA-NeXT-7B & 16.67 & 5921 & 4934 & 1156 & 487 & / & / \\
+Ours (11\%) & 2.68 & 2733 & 1769 & 191 & 153 & 3.87 & 48 \\
+Ours (22\%) & 4.11 & 3235 & 2540 & 313 & 181 & 4.47 & 57 \\
+Ours (33\%) & 5.61 & 3827 & 3207 & 437 & 214 & 5.15 & 66 \\
\bottomrule
\end{tabular}
}
\end{table*}
\section{Sensitivity of Pruning-Layer Choices}
\label{app:layer_sensitivity}

To further examine whether TRIO depends on a specific set of pruning-layer indices, we evaluate nearby pruning-layer configurations on LLaVA-1.5. As shown in Table~\ref{tab:layer_sensitivity}, shifting the pruning layers within a local range leads to only minor performance variations. This indicates that TRIO is stable within a reasonable neighborhood of layer choices, rather than being sensitive to one exact configuration.

\begin{table}[t]
\centering
\caption{Sensitivity of nearby pruning-layer choices on LLaVA-1.5.}
\label{tab:layer_sensitivity}
\begin{tabular}{lcc}
\toprule
Method & Pruning Layers & POPE \\
\midrule
TRIO (Ours) & $[1,10,15]$ & 84.3 \\
TRIO (Ours) & $[1,9,14]$  & 83.9 \\
TRIO (Ours) & $[1,11,16]$ & 84.2 \\
\bottomrule
\end{tabular}
\end{table}

\section{Theoretical Analysis}\label{Analysis}
\subsection{Per-layer FLOPs and Baseline}
We follow the FastV-style approximation and only count the dominant matrix-multiplication FLOPs
within each Transformer block (self-attention + FFN):
\begin{equation}
f(n)=4nd^{2}+2n^{2}d+2ndm.
\end{equation}
Without pruning, the token length remains $V_0$ for all layers, yielding the baseline FLOPs
\begin{equation}
F_{\mathrm{base}}=32f(V_0).
\end{equation}

\subsection{Backbone Inference FLOPs $F_{\mathrm{inf}}$ (Pruning at 1/10/15, Layers 1--32)}
Layers are indexed as $1,\ldots,32$. Pruning is performed at layers $1/10/15$ and becomes effective
from the next layer. Therefore, the four segments are:
(i) layer $1$: length $V_0$ (1 layer);
(ii) layers $2$--$10$: length $V_1$ (9 layers);
(iii) layers $11$--$15$: length $V_2$ (5 layers);
(iv) layers $16$--$32$: length $V_3$ (17 layers).
Hence,
\begin{equation}
F_{\mathrm{inf}}=f(V_0)+9f(V_1)+5f(V_2)+17f(V_3).
\end{equation}

\subsection{Local Gradient Overhead $F_{\mathrm{grad}}$}
At each pruning layer, we perform one extra forward pass of the current block to construct the proxy loss,
and one backward pass to obtain the token-level input gradients. Let $\beta$ denote the FLOPs ratio
of the backward pass to the forward pass under the same approximation. Then the overhead per pruning layer
is approximately
\begin{equation}
(1+\beta)f(V).
\end{equation}
Since the three pruning stages occur when the current visual token lengths are $V_0, V_1, V_2$ respectively,
we obtain
\begin{equation}
F_{\mathrm{grad}}=(1+\beta)\big(f(V_0)+f(V_1)+f(V_2)\big).
\end{equation}
Let $\gamma=1+\beta$, then
\begin{equation}
F_{\mathrm{grad}}=\gamma\big(f(V_0)+f(V_1)+f(V_2)\big).
\end{equation}


\subsection{Total FLOPs and Reduction Ratio}
The total overhead is
\begin{equation}
F_{\mathrm{ov}}=F_{\mathrm{grad}}+F_{\mathrm{nms}}
=\gamma\big(f(V_0)+f(V_1)+f(V_2)\big),
\end{equation}
and the total FLOPs after pruning is
\begin{equation}
F_{\mathrm{total}}=F_{\mathrm{inf}}+F_{\mathrm{ov}}.
\end{equation}
Therefore, the reduction ratio is
\begin{equation}
\mathrm{Saved}=1-\frac{F_{\mathrm{total}}}{F_{\mathrm{base}}}
=1-\frac{F_{\mathrm{inf}}+F_{\mathrm{ov}}}{32f(V_0)}.
\end{equation}

\section{Additional Efficiency Analysis}
\label{app:efficiency_analysis}

We provide additional efficiency analysis on LLaVA-NeXT-7B to further understand the runtime behavior of TRIO. Table~\ref{tab:overall_efficiency_local_timing} reports the overall efficiency and the local timing breakdown under different visual-token retention ratios. Compared with the full-token baseline, TRIO consistently reduces FLOPs, prefill time, and KV-cache memory. For example, under 11\% visual-token retention, TRIO reduces the average FLOPs from 16.67T to 2.68T, the prefill time from 4934s to 1769s, and the KV cache from 1156MB to 191MB.

Although TRIO introduces additional local overhead for proxy-loss computation and backward propagation, this overhead remains small compared with the overall runtime reduction. Under 11\% retention, the proxy-loss computation takes only 3.87ms, and the backward propagation takes 48ms in total. Meanwhile, the total runtime is reduced from 5921s to 2733s. This confirms that the cost of gradient-saliency estimation is effectively compensated by the sequence-length reduction in subsequent layers.

Table~\ref{tab:runtime_retention_ratio} further reports the measured total runtime under different visual-token retention ratios. As the retention ratio increases, more visual tokens are preserved, leading to higher runtime. Nevertheless, TRIO remains faster than the full-token baseline across a wide range of retention ratios. In particular, it achieves clear speedups from 11\% to 66\% retention. When the retention ratio reaches 77\%, the total runtime becomes slightly higher than the full-token baseline, indicating that the practical break-even point lies between 66\% and 77\%. This result suggests that TRIO is especially beneficial under moderate-to-aggressive visual-token compression settings.

\begin{table}[t]
\centering
\small
\setlength{\tabcolsep}{6pt}
\renewcommand{\arraystretch}{1.12}
\caption{
\textbf{Measured total runtime under different retention ratios on LLaVA-NeXT-7B.}
TRIO remains faster than the full-token baseline up to 66\% visual-token retention.
}
\label{tab:runtime_retention_ratio}
\begin{tabular}{lc}
\toprule
\textbf{Method} & \textbf{Total Time (s)} \\
\midrule
LLaVA-NeXT-7B & 5921 \\
+Ours (11\%) & 2733 \\
+Ours (22\%) & 3235 \\
+Ours (33\%) & 3827 \\
+Ours (44\%) & 4460 \\
+Ours (55\%) & 5012 \\
+Ours (66\%) & 5521 \\
+Ours (77\%) & 6137 \\
\bottomrule
\end{tabular}
\end{table}




\end{document}